\documentclass{article}
\usepackage[utf8]{inputenc}

\title{Histogram- and Diffusion-Based Medical Out-of-Distribution Detection}

\usepackage[numbers,sort]{natbib}
\usepackage{graphicx}
\usepackage{caption}
\usepackage{subcaption}
\usepackage{hyperref}
\usepackage[capitalise]{cleveref}
\usepackage{authblk}
\usepackage{soul}
\usepackage{color}
\usepackage{url}
\usepackage{blindtext}
\usepackage{multirow}
\usepackage{booktabs}
\usepackage{float}
\begin{document}

\author{Evi M.C. Huijben$^*$}
\author{Sina Amirrajab\thanks{E.M.C. Huijben  and S. Amirrajab contributed equally}}
\author{Josien P.W. Pluim}
\affil{Department of Biomedical Engineering, Eindhoven University of Technology, Eindhoven, The Netherlands}
\affil{\texttt{e.m.c.huijben@tue.nl - s.amirrajab@tue.nl}}

\date{}

\maketitle

\begin{abstract}

Out-of-distribution (OOD) detection is crucial for the safety and reliability of artificial intelligence algorithms, especially in the medical domain. In the context of the Medical OOD (MOOD) detection challenge 2023, we propose a pipeline that combines a histogram-based method and a diffusion-based method. The histogram-based method is designed to accurately detect homogeneous anomalies in the toy examples of the challenge, such as blobs with constant intensity values. The diffusion-based method is based on one of the latest methods for unsupervised anomaly detection, called DDPM-OOD. We explore this method and propose extensive post-processing steps for pixel-level and sample-level anomaly detection on brain MRI and abdominal CT data provided by the challenge. Our results show that the proposed DDPM method is sensitive to blur and bias field samples, but faces challenges with anatomical deformation, black slice, and swapped patches. These findings suggest that further research is needed to improve the performance of DDPM for OOD detection in medical images.


\end{abstract}

\section{Introduction}

Out-of-distribution (OOD) detection, the task of identifying inputs that do not belong to the data distribution on which the model was trained~\cite{yang2021generalized}, is crucial for ensuring the safety and reliability of machine learning systems in the medical domain, where unexpected inputs can lead to misdiagnosis and adverse outcomes. Generative models are powerful tools for learning the underlying distribution of complex medical data, which in turn can be used for OOD detection \cite{zhou2017anomaly, zimmerer2018context, chen2018unsupervised, marimont2021anomaly, wolleb2020descargan}.

Recently, diffusion models, introduced as a powerful category of generative models \cite{dhariwal2021diffusion, ho2020denoising}, have shown impressive results in generating realistic and diverse images from various domains in computer vision \cite{croitoru2023diffusion} and have found their way into unsupervised anomaly detection in medical images \cite{wyatt2022anoddpm, graham2023denoising, wolleb2022diffusion, pinaya2022fast, behrendt2023patched}.

To develop and benchmark methods for OOD detection, the Medical Out-of-Distribution Analysis Challenge (MOOD)\footnote{\url{http://medicalood.dkfz.de/web/}} \cite{zimmerer2022mood}, which includes sample-level and pixel-level anomaly detection for brain MRI and abdominal CT datasets, was first organized in conjunction with MICCAI 2020 and will be organized again in conjunction with MICCAI 2023.

\section{Dataset}
\label{sec:dataset}
The MOOD2023 dataset consists of 800 brain MRI scans ($256\times256\times256$) and 550 abdominal CT scans ($512\times512\times512$), which we randomly divided into a training (90\%) and validation (10\%) set. Additionally, a small toy set with one in-distribution sample and three samples with a homogeneous sphere inserted to represent an anomaly was provided for both regions.

To evaluate our model, we transformed 40 in-distribution validation set samples for both regions by applying the transformations described in Table~\ref{tab:transformations} four times.
\begin{table}[]
\footnotesize
\caption{Transformations used to create the OOD validation sets (based on \texttt{torchio.transforms}, except for RandomBlackSlice).}
\begin{tabular}{llll}
\toprule
\textbf{Transformation}      & \textbf{Parameters}   & \textbf{Value brain} & \textbf{Value abdomen} \\ 
\toprule
\multirow{2}{*}{RandomElasticDeformation} & max\_displacement & 30 & 50  \\ 
                                          & max\_displacement & 40 & 80  \\ \hline
\multirow{2}{*}{RandomBlur}               & std               & 2  & 3   \\ 
                                          & std               & 4  & 5   \\ \hline
\multirow{2}{*}{RandomBiasField}          & coefficients      & 1  & 1   \\ 
                                          & coefficients      & 2  & 2   \\ \hline
\multirow{2}{*}{RandomSwap (1 iteration)} & patch\_size       & 30 & 50  \\ 
                                          & patch\_size       & 80 & 100 \\ \hline
\multirow{2}{*}{RandomBlackSlice}         & slice\_thickness  & 1  & 1   \\ 
                                          & slice\_thickness  & 5  & 7   \\ 
\bottomrule
\end{tabular}
\label{tab:transformations}
\end{table}

\section{Method}
\begin{figure}
  \centering
  \includegraphics[width=\linewidth]{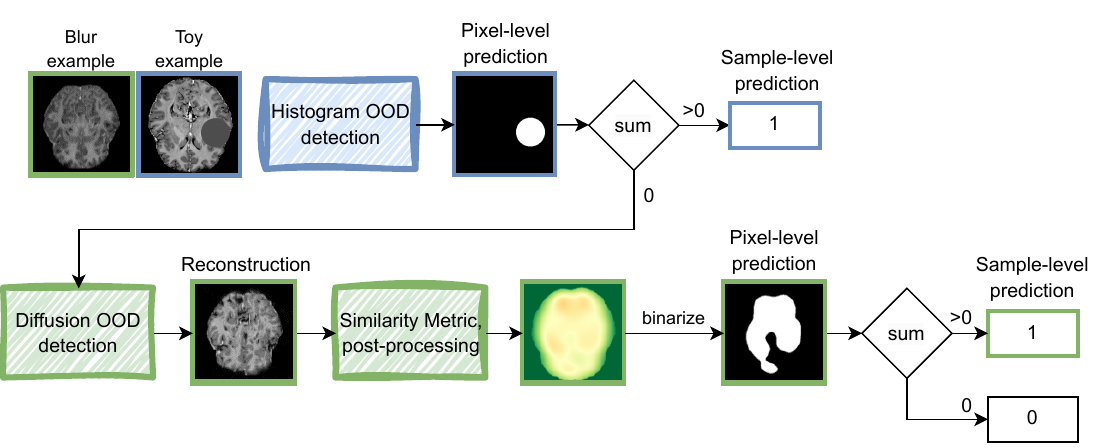}
  \caption{Anomaly detection pipeline using histogram OOD detection method (blue) and diffusion OOD detection method (green). The same pipeline is used for brain MRI and abdominal CT data.}
  \label{fig:pipeline}
\end{figure}

The proposed pipeline is shown in Figure~\ref{fig:pipeline}\footnote{Code available at \url{https://github.com/evihuijben/mood_challenge}}. In case of the abdominal dataset, the input images are first resized to $256\times256\times256$ before being processed through the pipeline. The histogram OOD detection module is then used for each input image to detect homogeneous anomalies, shown in blue. These anomalies are identified as specific intensity peaks in the image histogram. If no anomalies are detected by the histogram module, the image is passed on to the diffusion module. At this stage, the reconstruction error is measured using a pixel-wise structural similarity index measure (SSIM) map. The output is then post-processed to generate the final prediction, which is shown in green.


\subsection{Histogram OOD detection}

\begin{figure}
  \centering

  \includegraphics[width=\linewidth]{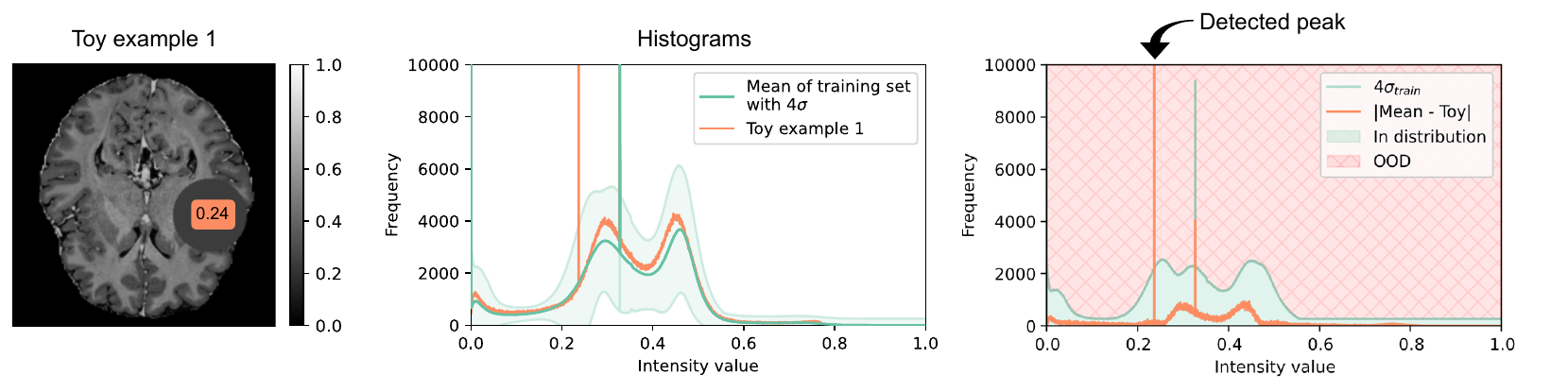}
  \caption{Histogram OOD detection to detect homogeneous regions in toy examples. All histograms are computed 3D images. We first subtract the histogram of the toy example from the mean histogram of the training data, and then detect the peak location (intensity value of 0.24) above four times the standard deviation.}  \label{fig:histogram brain}
\end{figure}

We propose a histogram-based OOD detection method that can detect only homogeneous anomalies, such as the toy examples in the challenge. We first compute the histogram mean and standard deviation (4096 bins) for the brain and abdomen based on the 3D training images. When calculating the histogram of an OOD test sample, homogeneous anomalies result in a peak with a much higher frequency than in the training set. As shown in Figure~\ref{fig:histogram brain}, the mean of the training set with four times the standard deviation is used to suppress the histogram of normal brain areas, leaving only the peak at intensity value 0.24 for toy example one. One spike appeared consistently for all brain cases, as seen in green in Figure~\ref{fig:histogram brain}. We found that this spike originated from homogeneous masking artifacts at the border of the applied brain mask. However, since this bin has a large standard deviation, it is never detected. In addition, we discard the zero intensity peak to avoid detection due to the large homogeneous background. Finally, we increase the standard deviations to 64 and 128 for brain and pelvis, respectively, to avoid false positives in intensities with little variation.
The final detected peak is used to threshold the image and create an anomaly mask. We further process the mask using erosion and dilation operations with a structure size of six to remove small erroneous objects.




\subsection{Diffusion OOD detection}
\subsubsection{Diffusion model}
Diffusion models are latent variable models, a variant of variational autoencoders, where the latent variable matches the dimension of the input data~\cite{ho2020denoising}. They include a forward diffusion process that incrementally adds noise to the data, and a backward diffusion process that uses deep learning-based denoising to generate realistic images.

Our diffusion-based OOD detection method is based on the work of Graham et al.~\cite{graham2023denoising}\footnote{\url{https://github.com/marksgraham/ddpm-ood}}, employing denoising diffusion probabilistic models (DDPMs)~\cite{ho2020denoising} for reconstruction-driven OOD detection. The key concept involves training the DDPM on the training data and then use it to reconstruct input images at test time. If the input image is within the training distribution, the model can reconstruct a high quality image. However, if the input image is OOD, the model yields a poor reconstruction, which is detectable by a reconstruction score. We trained a time-conditional 2D U-Net with attention mechanisms based on~\cite{rombach2022high,pinaya2022brain} to denoise images at each step in the backward process. Two identical models were trained independently for brain and abdomen, using the parameters described in Table~\ref{tab:implementation details}.

\begin{table}[ht]
\centering
\caption{Diffusion model parameters, built on top of \cite{graham2023denoising} and MONAI generative AI \cite{pinaya2023generative}.}
\begin{tabular}{ll}
\toprule

\hline
Noise & Simplex \cite{wyatt2022anoddpm}\\
Diffusion steps & 1000 \\
Scheduler & Scaled linear \cite{rombach2022high}\\
($\beta_{start}$,$\beta_{end}$)  & (0.001 , 0.015) \\ \hline
Optimizer & Adam \\
Learning rate & $2.5e-5$ \\
Loss & $MSE({noise}_{added}, {noise}_{predicted})$ \\
Batch size & 4 \\
Epoch & 60 \\
\hline
Intensity normalization & [0, 1] \\
Data augmentation & No \\
Input size (brain \& abdomen) & 2D axial slices $265 \times 256$ \\
\hline

\bottomrule
\end{tabular}
\label{tab:implementation details}
\end{table}




\subsubsection{Inference and post-processing}
Given the trained DDPM, we can perform multiple reconstructions at different diffusion steps, as suggested in \cite{graham2023denoising}, or determine the optimal step, as proposed in \cite{wyatt2022anoddpm}. Smaller steps (e.g., t=10 or 50) introduce minimal noise, resulting in anatomically similar reconstructions with faithful representation of anomalous regions. Conversely, larger steps (e.g., t=500 or 800) heavily distort the image content, causing anomalous regions to heal during reconstruction. Moreover, the reconstruction time becomes unacceptably long with larger steps. After experimental evaluation on our OOD validation set (see section~\ref{sec:dataset}), we chose t=200.

To obtain an OOD prediction score from the reconstruction, we first compute a body mask using \texttt{SimpleITK} algorithms. This involves Otsu thresholding on the input image, followed by binary dilation (radius = 5) and selection of the largest connected component. We then apply binary morphological closure (radius = 5) and hole filling to ensure complete object coverage. This body mask is then used to process the reconstruction image into a pixel-level OOD prediction. To achieve this, we compute a per-pixel SSIM map that compares the 3D masked reconstruction to the 3D input image. A 3-pixel border padding is added to the SSIM map to ensure accurate SSIM computation while minimizing border effects. Subsequently, we apply a Gaussian filter with a standard deviation of 15 to blur the SSIM map and threshold it at 0.5. The resulting binarized 3D map serves as the pixel-level prediction score for the MOOD challenge, with the sample-level OOD score set to 1 if at least one voxel is predicted to be OOD. The pixel-level predictions for the abdominal dataset are resized to the original resolution in the final post-processing step.

\begin{figure}[!ht]
  \centering

  \includegraphics[width=\linewidth]{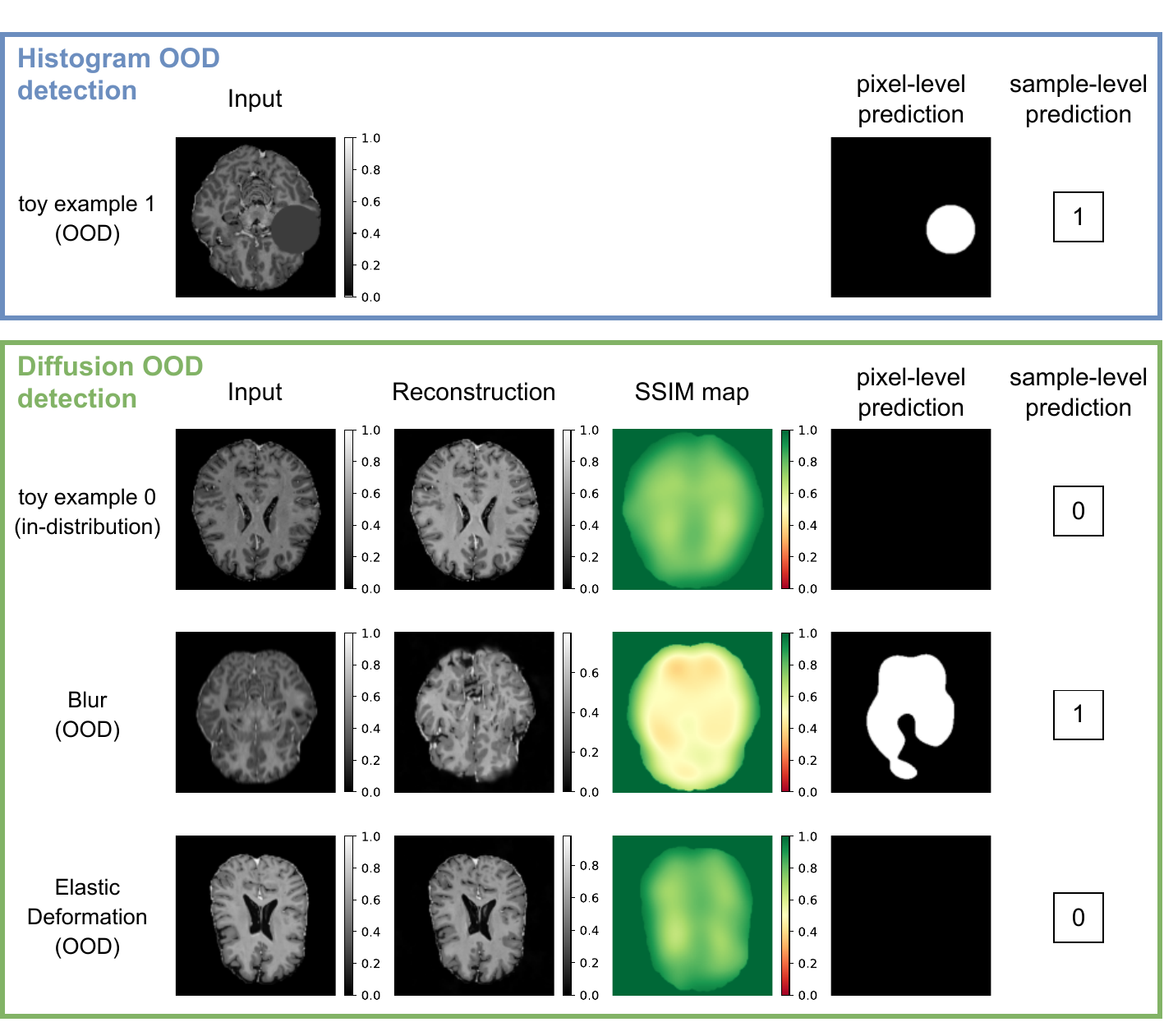}

  \caption{The qualitative results for pixel-level and sample-level predictions using our proposed histogram OOD detection (blue) and diffusion OOD detection (green).}
  \label{fig:results}
\end{figure}

\begin{figure}[!ht]
  \centering

  \includegraphics[width=\linewidth]{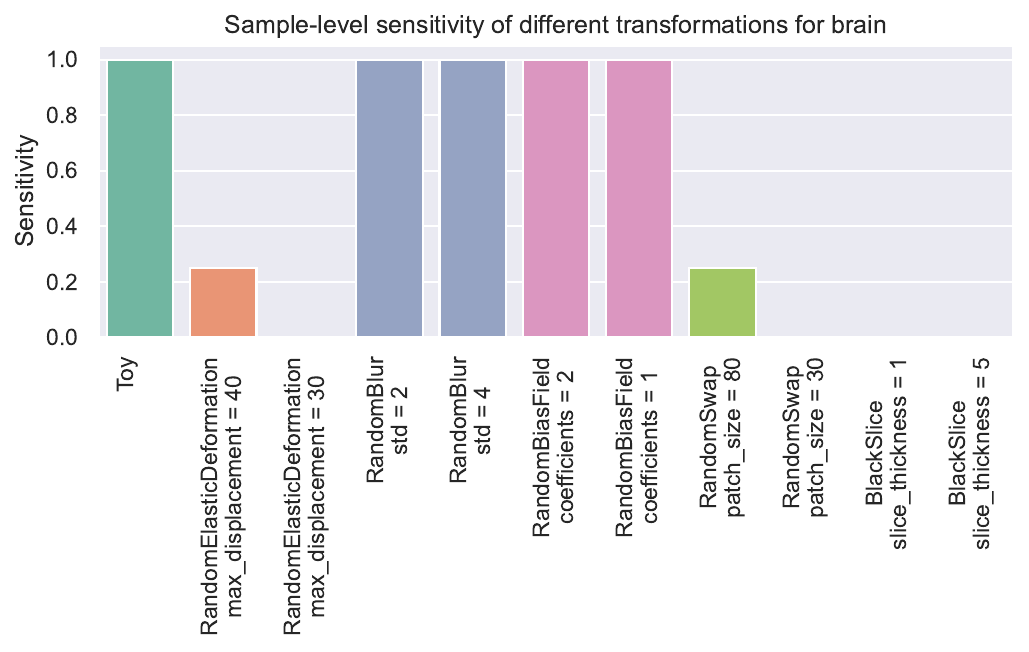}
  \caption{Sample-level sensitivity grouped by OOD type for brain cases.}
  \label{fig:barplot brain}
\end{figure}

\begin{figure}[!ht]
  \centering

  \includegraphics[width=\linewidth]{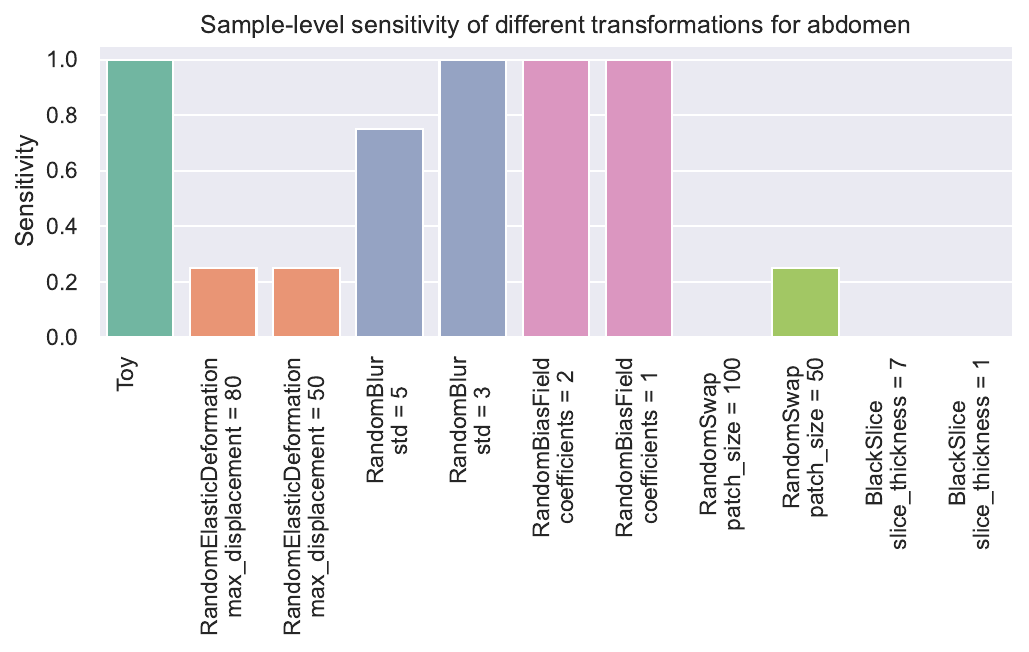}
  \caption{Sample-level sensitivity grouped by OOD type for abdomen cases.}
  \label{fig:barplot abdomen}
\end{figure}

\section{Results}
Figure~\ref{fig:results} shows qualitative results from a selection of the toy and OOD validation datasets for brain images. The first row (in blue) illustrates a toy example correctly identified by our histogram-based OOD detection. The last three rows (in green) depict one in-distribution and two OOD samples. As observed, the in-distribution sample is correctly classified, while the blurred example is accurately identified as OOD. However, our method fails to detect the elastic deformation. Figure~\ref{fig:barplot brain} presents the sample-level sensitivity categorized by different OOD groups, demonstrating that toy, blur, and bias field samples are easily detected, while deformation, swap, and black slice samples pose greater challenges. Furthermore, our method achieves a sample-level specificity of 1.00 for the in-distribution brain validation set. Similar patterns are observed for the abdomen dataset, both visually and quantitatively, as depicted in Figure~\ref{fig:barplot abdomen}, which shows the sensitivity of the OOD groups. Additionally, the sample-level specificity yields a similar value of 0.96.





\section{Discussion and Conclusion}
In this paper, we introduced a two-step approach for OOD detection in medical images: 1) histogram-based and 2) diffusion model-based. The histogram-based method excelled in identifying homogeneous anomalies with high accuracy, while the diffusion model-based approach detected OOD cases missed by the histogram-based method. However, we observed that the diffusion model lacked anatomical awareness, resulting in poor detection of cases involving deformations (e.g. elastic deformations or random patch swaps).

For future research, we recommend exploring anatomically aware models by using 2.5D or 3D models, and utilizing models that better capture spatial features, such as latent diffusion models~\cite{rombach2022high}. Additionally, since real-time OOD detection is essential, addressing the longer inference time inherent in diffusion models should be a priority for future work.

In conclusion, our proposed two-step OOD approach effectively identifies specific OOD cases in brain MRI and abdominal CT scans, which has the potential to enhance the application of machine learning in healthcare.

    

\bibliographystyle{plain}
\bibliography{bib}

\begin{thebibliography}{10}

\bibitem{behrendt2023patched}
Finn Behrendt, Debayan Bhattacharya, Julia Kr{\"u}ger, Roland Opfer, and
  Alexander Schlaefer.
\newblock Patched diffusion models for unsupervised anomaly detection in brain
  mri.
\newblock {\em arXiv preprint arXiv:2303.03758}, 2023.

\bibitem{chen2018unsupervised}
Xiaoran Chen and Ender Konukoglu.
\newblock Unsupervised detection of lesions in brain mri using constrained
  adversarial auto-encoders.
\newblock {\em arXiv preprint arXiv:1806.04972}, 2018.

\bibitem{croitoru2023diffusion}
Florinel-Alin Croitoru, Vlad Hondru, Radu~Tudor Ionescu, and Mubarak Shah.
\newblock Diffusion models in vision: A survey.
\newblock {\em IEEE Transactions on Pattern Analysis and Machine Intelligence},
  2023.

\bibitem{dhariwal2021diffusion}
Prafulla Dhariwal and Alexander Nichol.
\newblock Diffusion models beat gans on image synthesis.
\newblock {\em Advances in neural information processing systems},
  34:8780--8794, 2021.

\bibitem{graham2023denoising}
Mark~S Graham, Walter~HL Pinaya, Petru-Daniel Tudosiu, Parashkev Nachev,
  Sebastien Ourselin, and Jorge Cardoso.
\newblock Denoising diffusion models for out-of-distribution detection.
\newblock In {\em Proceedings of the IEEE/CVF Conference on Computer Vision and
  Pattern Recognition}, pages 2947--2956, 2023.

\bibitem{ho2020denoising}
Jonathan Ho, Ajay Jain, and Pieter Abbeel.
\newblock Denoising diffusion probabilistic models.
\newblock {\em Advances in neural information processing systems},
  33:6840--6851, 2020.

\bibitem{marimont2021anomaly}
Sergio~Naval Marimont and Giacomo Tarroni.
\newblock Anomaly detection through latent space restoration using vector
  quantized variational autoencoders.
\newblock In {\em 2021 IEEE 18th International Symposium on Biomedical Imaging
  (ISBI)}, pages 1764--1767. IEEE, 2021.

\bibitem{pinaya2022fast}
Walter~HL Pinaya, Mark~S Graham, Robert Gray, Pedro~F Da~Costa, Petru-Daniel
  Tudosiu, Paul Wright, Yee~H Mah, Andrew~D MacKinnon, James~T Teo, Rolf Jager,
  et~al.
\newblock Fast unsupervised brain anomaly detection and segmentation with
  diffusion models.
\newblock In {\em International Conference on Medical Image Computing and
  Computer-Assisted Intervention}, pages 705--714. Springer, 2022.

\bibitem{pinaya2023generative}
Walter~HL Pinaya, Mark~S Graham, Eric Kerfoot, Petru-Daniel Tudosiu, Jessica
  Dafflon, Virginia Fernandez, Pedro Sanchez, Julia Wolleb, Pedro~F da~Costa,
  Ashay Patel, et~al.
\newblock Generative ai for medical imaging: extending the monai framework.
\newblock {\em arXiv preprint arXiv:2307.15208}, 2023.

\bibitem{pinaya2022brain}
Walter~HL Pinaya, Petru-Daniel Tudosiu, Jessica Dafflon, Pedro~F Da~Costa,
  Virginia Fernandez, Parashkev Nachev, Sebastien Ourselin, and M~Jorge
  Cardoso.
\newblock Brain imaging generation with latent diffusion models.
\newblock In {\em MICCAI Workshop on Deep Generative Models}, pages 117--126.
  Springer, 2022.

\bibitem{rombach2022high}
Robin Rombach, Andreas Blattmann, Dominik Lorenz, Patrick Esser, and Bj{\"o}rn
  Ommer.
\newblock High-resolution image synthesis with latent diffusion models.
\newblock In {\em Proceedings of the IEEE/CVF conference on computer vision and
  pattern recognition}, pages 10684--10695, 2022.

\bibitem{wolleb2022diffusion}
Julia Wolleb, Florentin Bieder, Robin Sandk{\"u}hler, and Philippe~C Cattin.
\newblock Diffusion models for medical anomaly detection.
\newblock In {\em International Conference on Medical image computing and
  computer-assisted intervention}, pages 35--45. Springer, 2022.

\bibitem{wolleb2020descargan}
Julia Wolleb, Robin Sandk{\"u}hler, and Philippe~C Cattin.
\newblock Descargan: Disease-specific anomaly detection with weak supervision.
\newblock In {\em Medical Image Computing and Computer Assisted
  Intervention--MICCAI 2020: 23rd International Conference, Lima, Peru, October
  4--8, 2020, Proceedings, Part IV 23}, pages 14--24. Springer, 2020.

\bibitem{wyatt2022anoddpm}
Julian Wyatt, Adam Leach, Sebastian~M Schmon, and Chris~G Willcocks.
\newblock Anoddpm: Anomaly detection with denoising diffusion probabilistic
  models using simplex noise.
\newblock In {\em Proceedings of the IEEE/CVF Conference on Computer Vision and
  Pattern Recognition}, pages 650--656, 2022.

\bibitem{yang2021generalized}
Jingkang Yang, Kaiyang Zhou, Yixuan Li, and Ziwei Liu.
\newblock Generalized out-of-distribution detection: A survey.
\newblock {\em arXiv preprint arXiv:2110.11334}, 2021.

\bibitem{zhou2017anomaly}
Chong Zhou and Randy~C Paffenroth.
\newblock Anomaly detection with robust deep autoencoders.
\newblock In {\em Proceedings of the 23rd ACM SIGKDD international conference
  on knowledge discovery and data mining}, pages 665--674, 2017.

\bibitem{zimmerer2022mood}
David Zimmerer, Peter~M Full, Fabian Isensee, Paul J{\"a}ger, Tim Adler, Jens
  Petersen, Gregor K{\"o}hler, Tobias Ross, Annika Reinke, Antanas Kascenas,
  et~al.
\newblock Mood 2020: A public benchmark for out-of-distribution detection and
  localization on medical images.
\newblock {\em IEEE Transactions on Medical Imaging}, 41(10):2728--2738, 2022.

\bibitem{zimmerer2018context}
David Zimmerer, Simon~AA Kohl, Jens Petersen, Fabian Isensee, and Klaus~H
  Maier-Hein.
\newblock Context-encoding variational autoencoder for unsupervised anomaly
  detection.
\newblock {\em arXiv preprint arXiv:1812.05941}, 2018.

\end{thebibliography}

\end{document}